\pdfoutput=1

\documentclass[11pt]{article}
\usepackage{authblk}

\usepackage[dvipsnames]{xcolor}

\usepackage[]{acl}

\usepackage{times}
\usepackage{latexsym}

\usepackage[T1]{fontenc}

\usepackage[utf8]{inputenc}

\usepackage{booktabs}
\usepackage{tabularx}
\usepackage{lipsum}
\usepackage{graphicx}
\usepackage{todonotes}
\usepackage{svg}
\usepackage{amsmath}
\usepackage{multirow}
\usepackage{csquotes}
\usepackage{lingmacros}
\usepackage{float}
\usepackage{caption}
\usepackage{subcaption}

\usepackage{microtype}

\title{ZS4IE: A toolkit for Zero-Shot Information Extraction \\ with simple Verbalizations}

\author[1,*]{\textbf{Oscar Sainz}}
\author[2,*]{\textbf{Haoling Qiu}}
\author[1]{\\\textbf{Oier Lopez de Lacalle}}
\author[1]{\textbf{Eneko Agirre}}
\author[2]{\textbf{Bonan Min}}

\affil[1]{HiTZ Basque Center for Language Technologies - Ixa NLP Group}
\affil[ ]{University of the Basque Country (UPV/EHU)}
\affil[2]{Raytheon BBN Technologies}
\affil[ ]{ oscar.sainz@ehu.eus, haoling.qiu@raytheon.com}

\date{}

\begin{document}
\maketitle

\def\thefootnote{*}\footnotetext{Denotes equal contribution.}\def\thefootnote{\arabic{footnote}}

\begin{abstract}
The current workflow for Information Extraction (IE) analysts involves the definition of the entities/relations of interest and a training corpus with annotated examples. In this demonstration we introduce a new workflow where the analyst directly verbalizes the entities/relations, which are then used by a Textual Entailment model to perform zero-shot IE. We present the design and implementation of a toolkit with a user interface, as well as experiments on four IE tasks that show that the system achieves very good performance at zero-shot learning using only 5--15 minutes per type of a user's effort. Our demonstration system is open-sourced at \url{https://github.com/BBN-E/ZS4IE}. A demonstration video is available at \url{https://vimeo.com/676138340}.
\end{abstract}

\section{Introduction}


Information Extraction (IE) systems are very costly to build. The current \textbf{define-then-annotate-and-train} workflow uses supervised machine learning, where the analyst first defines the schema with the entities and relations of interest and then builds a training corpus with annotated examples. Unfortunately, each new domain and schema requires starting from scratch, as there is very little transfer between domains. 

\begin{figure*}
    \centering
    \resizebox{1.\textwidth}{!}{
        \includegraphics{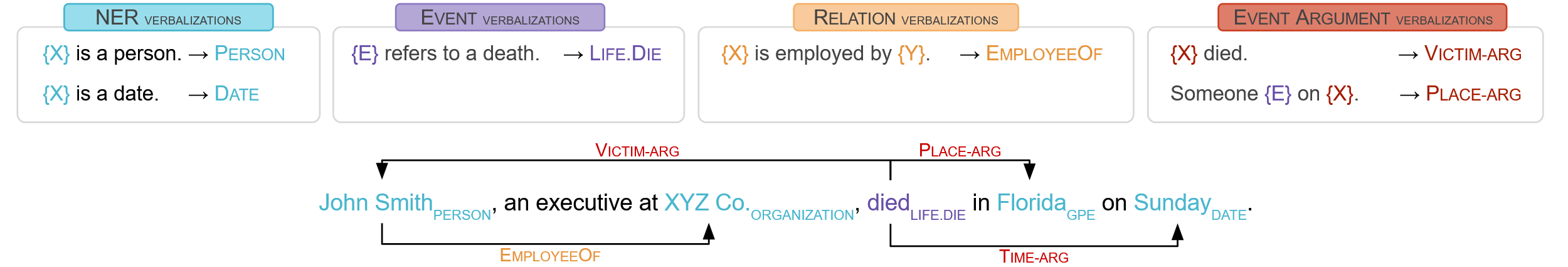}
    }
    \caption{Verbalization templates for a sample schema involving four tasks (from left to right, NER, EE, RE, EAE), with example output (bottom). The schema contains a {\color{BurntOrange} \textsc{EmployeeOf} relation} between {\color{Aquamarine} \textsc{Person}  } and {\color{Aquamarine} \textsc{Organization} entities} and a {\color{BlueViolet} \textsc{Life.Die} event} with three {\color{BrickRed} argument types} ({\color{BrickRed} \textsc{Victim}, \textsc{Place} and \textsc{Time}}) and  {\color{Aquamarine} \textsc{Person}, \textsc{Date} } and {\color{Aquamarine} \textsc{GPE}  entities} as fillers. Due to space constraints, at most two verbalizations per task shown.}
    \label{fig:verbalizations}
\end{figure*}

We present an alternative \textbf{verbalize-while-defining workflow} where the analyst defines the schema interactively in a user interface using natural language verbalizations of the target entity and relation types. Figure \ref{fig:verbalizations} shows sample verbalization templates for a simple schema involving an employee relation and a passing away event, as well as a sample output annotated with the schema. The annotation of the \textsc{EmployeeOf} relation requires performing Named Entity Recognition (NER)~\cite{tjong-kim-sang-de-meulder-2003-introduction} and Relation Extraction (RE)~\cite{zhang-etal-2017-position}, while annotating the \textsc{Life.Die} event involves NER, Event Extraction (EE), and Event Argument Extraction (EAE)~\cite{ACE}.  Our toolkit is able to perform those four IE tasks using a single user interface, allowing the analyst to easily model and test the schema without the need to annotate examples.

Our toolkit leans on recent work which has successfully recast several IE tasks  as Textual Entailment (TE) tasks \cite{white-etal-2017-inference, DBLP:journals/corr/abs-1804-08207, levy-etal-2017-zero, sainz-etal-2021-label}. For instance, \citet{sainz-etal-2021-label} model relation types between entity pairs using type-specific verbalization templates that describe the relation, generates a verbalization (hypothesis) automatically using those templates and then uses a pre-trained TE model to predict if the premise (the sentence where the pair appears) entails the hypothesis, therefore leading to a prediction of the relation or ``no relation''.



In this paper we thus present ZS4IE, a toolkit for zero-shot IE. We show that the four mainstream IE tasks mentioned above can be reformulated as TE problems, and that it is possible to achieve strong zero-shot performances leveraging pre-trained TE models and a small amount of templates curated by the user. Our toolkit allows a novice user to curate templates for each new types of entities, relations, events, and event argument roles, and validate their effectiveness online over any example. We also present strong results on widely used datasets with only 5-15 minutes per type of a user's effort. 

\begin{figure*}
    \centering
    \resizebox{1.\textwidth}{!}{
        \includegraphics{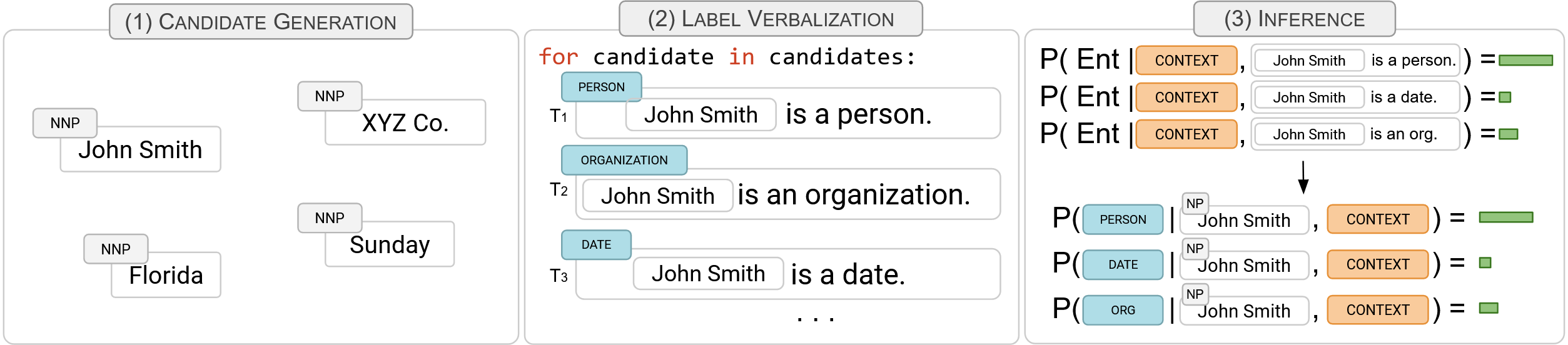}
    }
    \caption{Three steps for entailment-based NER. The steps for the other IE tasks is analogous. 
    }
    \label{fig:overall_illustration}
    \vspace{-.75em}
\end{figure*}

\section{Related Work} \label{sec:related_work}

Textual Entailment has been shown to be a reasonable proxy for classification tasks like topic or sentiment analysis \cite{yin-etal-2019-benchmarking,sainz-rigau-2021-ask2transformers,zhong-etal-2021-adapting-language}. To reformulate a classification problem as TE, it often starts with defining templates to describe each class label, leading to a natural language text (a ``verbalization'' of a hypothesis) for each possible label. Inference is performed by selecting the most probable candidate hypothesis entailing the premise. TE is usually implemented with pre-trained language model fine-tuned on TE datasets, such as MNLI \cite{williams-etal-2018-broad}, SNLI \cite{bowman-etal-2015-large}, FEVER \cite{thorne-etal-2018-fever}, ANLI \cite{nie-etal-2020-adversarial} or XNLI \cite{conneau-etal-2018-xnli}. The results on classification have been particularly strong for zero-shot and few-shot learning, with \citet{wang2021entailment} hypothesizing that entailment is a true language understanding task, where a model that performs entailment well is likely to succeed on similarly-framed tasks.

\citet{sainz-etal-2021-label} reformulated relation extraction as a TE task surpassing the state-of-the-art in zero- and few-short learning. A similar approach was previously explored by \citet{obamuyide-vlachos-2018-zero}, using TE models that are not based on pre-trained language models. Similar to TE, \cite{clark-etal-2019-boolq} performs yes/no Question Answering, in which a model is asked about the veracity of some fact given a passage. \citet{lyu-etal-2021-zero} recast the zero-shot event extraction as a TE task, using TE model to check whether a piece of text is about a type of event. Lastly, \citet{sainz-etal-2022-textual} showed that TE allows to leverage the knowledge from other tasks and schemas.

\section{IE via Textual Entailment}

We first describe how to recast each of the IE tasks (NER, RE, EE, EAE) as TE independently, and leave the workflow between the tasks for the next section.
At a high level, the zero-shot TE reformulation consists of three steps: candidate generation, label verbalization and TE inference (Figure~\ref{fig:overall_illustration} illustrates the steps for NER). The first step, candidate generation, identifies text spans (e.g., proper nouns for NER) or span pairs (a pair of entity mentions for  relation extraction) in the input sentence as the focus of the prediction. Taking a text span (or span pair) as input, the label verbalization step applies a verbalization template to generate a {\it hypothesis}, which is a natural language sentence describing the span (or span pair) being an instance of a type of entity, relation, event, or event argument. The verbalization generates hypothesis for each of the target types. Finally, the TE inference step takes the original sentence (the {\it premise}) and each {\it hypothesis} as input, and uses a pre-trained TE model to predict if the {\it premise} entails, contradicts, or is neutral to the {\it hypothesis}. The type with the verbalization having the highest entailment probability is selected. We next describe each step in detail. 





\subsection{Candidate Generation} \label{sec:candidate-gen}

We describe the candidate generation for each of the task below. 

\paragraph{Named Entity Recognition (NER):} Candidates are extracted using specific patterns of PoS tags as returned by Stanza~\cite{qi-etal-2020-stanza}. For instance, for the simple example in Figure~\ref{fig:verbalizations} it suffices to select proper nouns (shown in Figure~\ref{fig:overall_illustration}), which are easily extended with other PoS patterns if needed. The toolkit also allows the usage of a constituency parser~\cite{kitaev-klein-2018-constituency}.  

\paragraph{Relation Extraction (RE):} Each relation requires a pair of entities that satisfy specific type constraints, e.g. the \textsc{EmployeeOf} relation requires a \textsc{Person} and an \textsc{Organization}. A NER module is used to extract all candidate entities that follow the required entity types according to the target schema. The toolkit uses the TE based NER module, although it also allows usage of a supervised NER system ~\cite{qi-etal-2020-stanza}. 

\paragraph{Event (Trigger) Extraction (EE):} The main goal of this task is to detect whether the input sentence contains a mention of any of the target event types in the schema, e.g. \textsc{Life.Die}. This task can be formulated as a multi-label text classification task, and in this case the full sentence is the candidate. Alternatively, the textual span that most likely expresses the event (the so-called trigger) can be extracted. In this case, the candidates are generated using specific PoS tags, e.g. verbs like {\it died} (cf. Figure ~\ref{fig:verbalizations}). Our toolkit allows both options. 


\paragraph{Event Argument Extraction:} Given a sentence containing an event type (as detected by EE above), the goal is to extract entity mentions that are fillers of the target arguments in the schema. For example, the schema in Figure ~\ref{fig:verbalizations} involves three target arguments. Each of the arguments requires specific entity types, e.g. \textsc{Person} for the \textsc{Victim} argument. The candidates of the required types are extracted using the same NER module as for RE. 

\subsection{Label Verbalization}

For each of the IE tasks, the label verbalization process takes a sentence, a set of candidates and the set of target types (e.g. NER types), and generates a natural language text (the hypothesis) describing the existence of the type in the sentence (the premise) using verbalization templates. Each candidate is a span (or pair of spans) that can belong to a specific type (e.g. being a \textsc{Person} in NER). Therefore, the textual verbalization is generated to express each potential type for the span or the pair of spans. For the NER and event extraction tasks, each verbalization expresses one potential entity (or event type) for the target candidate. For the relation and event argument extraction tasks, the verbalization template combines the information from the text spans of the candidate pair and produces a text that expresses a relation (or event argument role). The analyst just needs to write the verbalization templates for each target type, and they are applied to the candidates to generate the hypothesis, as shown in the second step in Figure~\ref{fig:overall_illustration} for NER. 




Figure~\ref{fig:verbalizations} shows sample TE verbalization templates for entity, relation, event, and event argument types corresponding to the 4 IE tasks, as well as sample example as output. The templates for \textbf{NER} and \textbf{event extraction} (leftmost part of the figure) are applied over a single candidate as extracted in the previous step (the candidate entity or event trigger, respectively). Note that for event extraction it is also possible to produce hypothesis using templates with no slots, e.g. "A person died" for \textsc{Life.Die}. In the case of \textbf{relation extraction}, the verbalization templates contain two slots for the two entity spans potentially holding the relation. Finally, templates for \textbf{event argument extraction} can be more varied. The figure shows two examples: a template using a single slot for the candidate filler, and a template which, in addition to the filler slot, uses the trigger ("died" in this case, for \textsc{Place}).

\subsection{Inference}

Given a premise (the original sentence) and a hypothesis (an verbalization generated by label verbalization templates), we use a pre-trained TE model to decide whether the hypothesis is entailed by, contradicted with, or is neutral to the premise. In principle, any model trained on an entailment dataset can be used. 
%
The inference is mainly determined by three key factors: the TE probabilities for the verbalizations of all templates for all labels, the type-specific input span constraints, and a threshold that decides if the probability is high enough to consider the candidate a positive instance. The type-specific input span constraints are enforced to make sure we don't have candidates that violates the constraints. We return the class label of the hypothesis with highest entailment probability. If none of the hypothesis is higher than the threshold, we return the negative class, that is the class that represents that there is not a valid entity, relation, event, or event argument role type for the input candidate. The threshold for minimal entailment probability is set by default to 0.5.

\section{ZS4IE toolkit}

ZS4IE comprises a pipeline and a user interface.

\subsection{The ZS4IE Pipeline} \label{sec:pipeline}
\begin{figure}
    \centering
    \resizebox{0.48\textwidth}{!}{
        \includegraphics{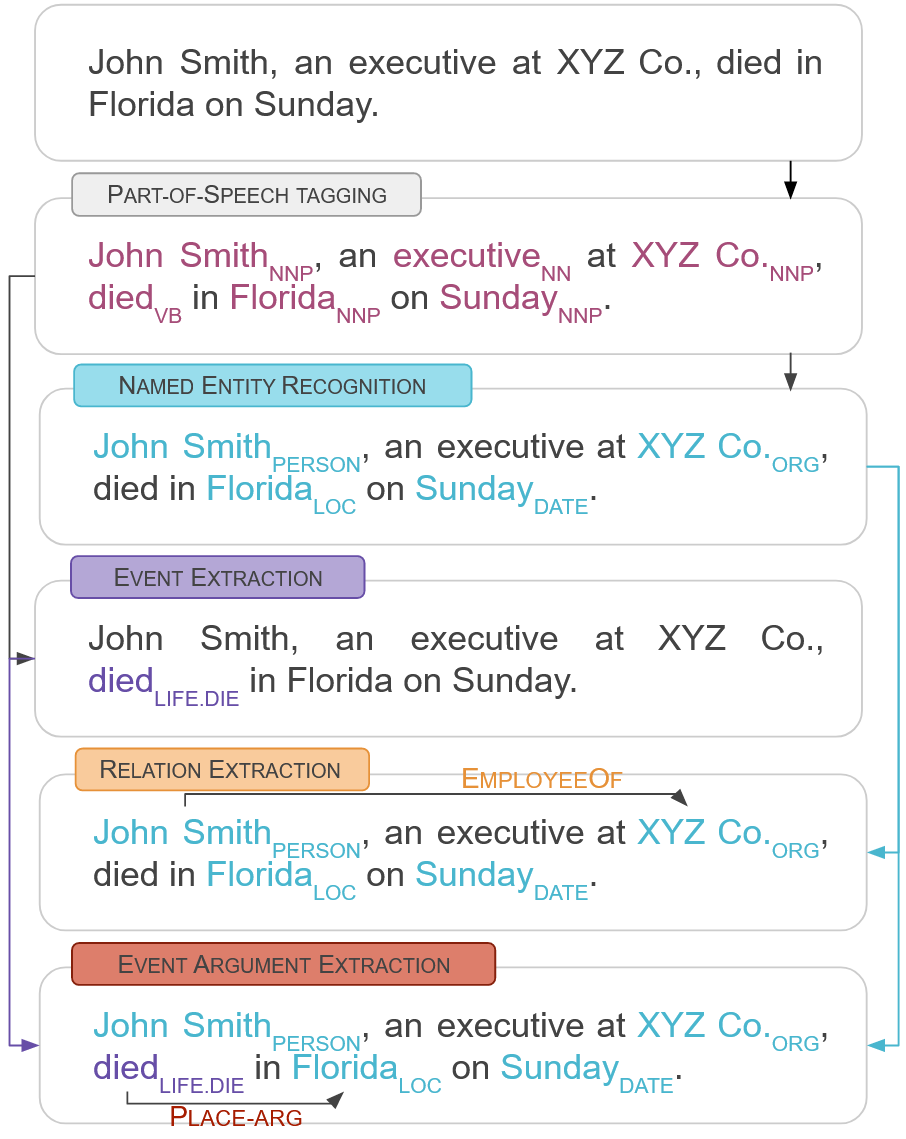}
    }
    \caption{An illustration of the dependencies between the four IE tasks.}
    \label{fig:overall_arch}
\end{figure}

As described in Section~\ref{sec:candidate-gen} and illustrated in Figure ~\ref{fig:overall_arch}, there are inter-task dependencies between the four IE tasks (e.g., relation extraction requires that entity mentions have already been tagged in the input sentence). Some task also require external NLP tools for generating candidates. To address these issues and to allow maximal flexibility for the users, we support the following two workflows. 

\paragraph{The End-to-End (E2E) Mode:} This mode will run the ZS4IE modules in a pipeline: we allow the users to start from raw text, and perform customization (e.g., develop templates for new types of interest) for all four IE tasks. The user has to follow the inter-task dependencies as illustrated in Figure~\ref{fig:overall_arch}: the user must finish NER customization before moving on to relation extraction or the event argument extraction task, because the later two tasks needs NER to generate their input candidates. Similarly, the user must finish customization for the event trigger classification task, before working on the event argument extraction task. 

The end-to-end pipeline also runs a customizable pre-processing step including a POS tagger and a constituency parser, before any of the later modules. 

\paragraph{The Task Mode: } In this mode, the user can choose to work on each of the four IE tasks independently.  In order to address the inter-dependencies, the user can choose to run an independent NER module instead, as part of the pre-processing step. 
The user interface allows the user to tag any spans for entity or event trigger types, before running customization for the more complex tasks such as relation extraction or event argument extraction. This option allows to explore additional entity and event trigger types before actually implementing them

\subsection{User Interface (UI)}

\begin{figure}
    \centering
    \resizebox{\linewidth}{!}{
        \includegraphics{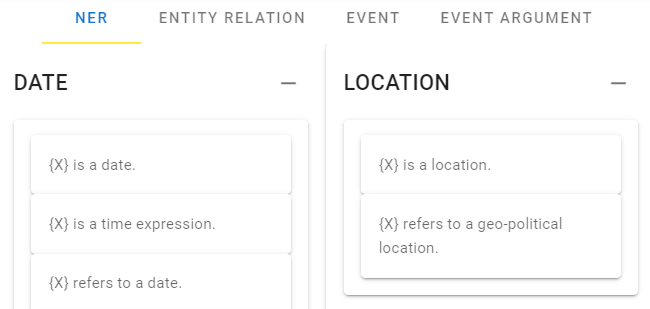}
    }
    \caption{The UI for curating templates for types of interests for NER, relation extraction, event extraction and event argument extraction tasks. The NER tab is partially shown with two types. }
    \label{fig:ui1}
\end{figure}
\begin{figure}
    \centering
    \resizebox{\linewidth}{!}{
         \includegraphics{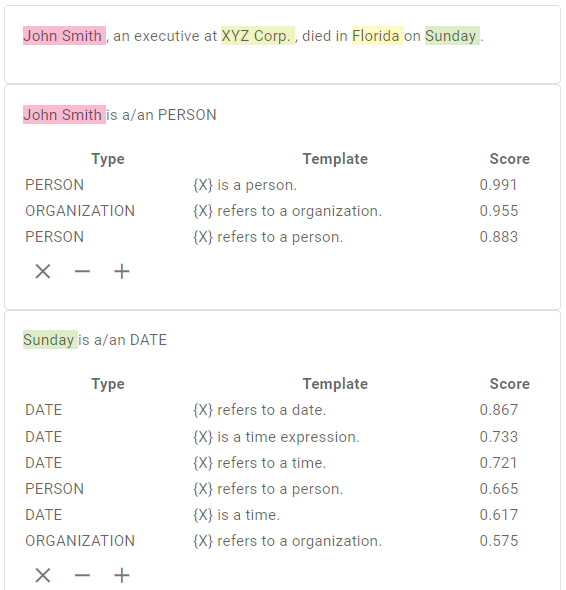}
    }
    \caption{The UI for displaying NER extraction results on an user-input sentence. We show the extractions and the likelihood scores of the templates above the threshold (e.g. $\mathcal{T}=0.5$).}
    \label{fig:ui2}
\end{figure}

Figure~\ref{fig:ui1} shows the User Interface. It allows the user to add new types of entities, relations, events and event argument roles, and then develop templates (along with input type constraints for each type). Figure~\ref{fig:ui2} shows the NER extraction results on an user-input sentence. It also displays the likelihood scores produced by the TE model of those templates that are above the threshold, to allow the user to validate templates.

To show why it extracts each entity, it displays a ranked list of likely entity types, the template that led to that type, along with the entailment probability produced by the pre-trained TE model. The user can click on "+" and "-" sign next to each extraction to label its correctness. Our system will track the total number of extractions and and accuracy for each task, each type and each template, to allow the user to quickly validate the effectiveness of the templates and to spot any low-precision template.

\paragraph{Supplying Input Text: } The user can supply a text snippet, one at a time, to test writing templates. As described in Section~\ref{sec:pipeline}, when using the {\t task} mode, the user can label spans in the input text for the more complex relation extraction and event argument extraction tasks, so that the text already has the right entity or event trigger spans and types to begin with.

\paragraph{Develop Templates for New Types:} The user can add new types of entities, relations, events, and event argument role. For each type, the user can create templates along with the input span type constraints, and then run inference interactively on the input text, to see whether these templates can be used for extract the instances. The user can label the correctness of the extracted instances, resulting a small development dataset (the {\it dev} set) to help measuring the precision and relative recall for each template, and to tune the threshold for the TE inference.

\paragraph{Display Metrics: } The UI displays the accuracy and yield for each template and each type in real-time, to allow the user to monitor the progress and make adjustments on the fly. 

More screenshots and details of our UI are describe in Appendix~\ref{sec:ui}.

\section{Experiments}


\begin{table*}[!ht]
    \centering
    \resizebox{\linewidth}{!}{
    \begin{tabular}{l|rrr|ccc|ccc|ccc|c}
        \toprule
         & \multicolumn{3}{c|}{NER} & \multicolumn{3}{c|}{RE} & \multicolumn{3}{c|}{EE} & \multicolumn{3}{c|}{EAE} & AVG \\
        Model & Pre & Rec & F1 & Prec & Rec & F1 & Prec & Rec & F1 & Prec & Rec & F1 & F1 \\
        \midrule
        
        
        
        RoBERTa & 53.3 & 54.5 & 53.9 & 32.8 & 75.5 & 45.7 & \textbf{23.8} & \textbf{63.0} & \textbf{34.5} & 20.5 & 60.9 & 30.7 & 46.7 \\
        
        RoBERTa* & \textbf{73.5} & \textbf{76.3} & \textbf{74.9} & 36.8 & 76.7 & 49.8 & 23.5 & 60.8 & 33.9 & \textbf{30.1} & \textbf{63.2} & \textbf{40.8} & \textbf{49.0} \\
        
        DeBERTa & 58.0 & 50.2 & 53.8 & \textbf{40.3} & \textbf{77.7} & \textbf{53.0} & 12.9 & 60.3 & 21.2 & 20.0 & 31.9 & 24.6 & 45.1 \\
        
        \midrule
        
        RoBERTa ({\small + $\mathcal{T}$ opt}) & \small{49.3} & \small{61.8} & \small{\textcolor{OliveGreen}{$\uparrow$} 54.9} & \small{56.1} & \small{55.8} & \small{\textcolor{OliveGreen}{$\uparrow$} 55.9} & \small{32.0} & \small{52.9} & \small{\textcolor{OliveGreen}{$\uparrow$} 39.9} & \small{25.8} & \small{40.1} & \small{\textcolor{OliveGreen}{$\uparrow$} 31.4} & \small{\textcolor{OliveGreen}{$\uparrow$} 50.9} \\
        RoBERTa* ({\small + $\mathcal{T}$ opt}) & \small{71.9} & \small{77.8} & \small{\textcolor{BrickRed}{$\downarrow$} 74.8} & \small{54.2} & \small{59.5} & \small{\textcolor{OliveGreen}{$\uparrow$} 56.8} & \small{25.1} & \small{58.6} & \small{\textcolor{OliveGreen}{$\uparrow$} 35.1} & \small{31.1} & \small{58.3} & \small{\textcolor{BrickRed}{$\downarrow$} 40.6} & \small{\textcolor{OliveGreen}{$\uparrow$} 51.9} \\
        DeBERTa ({\small + $\mathcal{T}$ opt}) & \small{56.3} & \small{63.1} & \small{ \textcolor{OliveGreen}{$\uparrow$} 59.5} & \small{66.3} & \small{59.7} & \small{\textcolor{OliveGreen}{$\uparrow$} 62.8} & \small{13.0} & \small{55.8} & \small{\textcolor{BrickRed}{$\downarrow$} 21.1} & \small{28.9} & \small{17.5} & \small{\textcolor{BrickRed}{$\downarrow$} 21.8} & \small{\textcolor{OliveGreen}{$\uparrow$} 51.3} \\
        \midrule
        Other authors & - & - & - & - & - & 49.2 & 36.2$\dagger$ & 69.1$\dagger$ & 47.5$\dagger$ & 38.2 & 35.8 & 37.0 & - \\
        
        \bottomrule
    \end{tabular}
    }
    \caption{Results for NER, RE, EE and EAE experiments results. 
    Three top rows for zero-shot systems with default parameters. Middle rows for threshold optimized on development. 
    The best scores among our results obtained with default thresholds are marked in \textbf{bold}. The $\dagger$ indicates non-comparable results due to additional SRL preprocessing.}
    \label{tab:results}
    \vspace{-1em}
\end{table*}

We evaluated our system using publicly available datasets. We use CoNLL 2003~\cite{tjong-kim-sang-de-meulder-2003-introduction} for NER evaluation, TACRED~\cite{zhang-etal-2017-position} for RE, and ACE for EE and EAE~\cite{ACE}. We evaluate each task independently (not as a pipeline) to make as comparable as possible to existing zero-shot systems. In order to apply our toolkit we made some adaptations as follows: We consider only proper nouns as candidates for NER, and we ignore  
the \textsc{Misc} label because it is not properly defined in the task~\footnote{More specifically, we re-labeled the \textsc{Misc} instances to \textsc{O} label.}. We evaluate EE as event classification, where the task is to output the events mentioned in the sentence without extracting the trigger words, as we found that deciding which is the trigger word is in many cases an arbitrary decision~\footnote{Note that EAE  can be addressed without an explicit mention of the trigger since we used templates that do not require the trigger}. 
In the case of RE we used the templates from \cite{sainz-etal-2021-label}, which are publicly available. We will release the templates used on the experiments as additional material along with the paper. The analysts spent between 5-15 minutes per type, depending on the task, with NER and EE being the fastest.

Table~\ref{tab:results} shows the zero-shot results for NER, RE, EE, and EAE tasks.  
We report the results of three entailment models: RoBERTa~\cite{roberta} trained on MNLI, RoBERTa* trained on MNLI, SNLI, FEVER and ANLI; and DeBERTa~\cite{he2021deberta} trained on MNLI.
The main results (top three rows) use the default threshold ($\mathcal{T}=0.5$), we selected the $\mathcal{T}$ blindly, without checking any development result.

The results show strong zero-shot performance. Note that there is no best entailment model, suggesting that there still exists margin for improvement. However, we see that RoBERTa* performs relatively well in all scenarios except EE (see Section \ref{sec:discussion} for further discussion). 

The table also shows in the middle three rows the results where we optimize the threshold on development. The results improve in most of the cases, and allow comparison to other zero-shot systems which sometimes optimize a threshold in development data.  

Furthermore, we compare our system with zero-shot task specific approaches from other authors when available. For RE, \citet{wang-etal-2021-zero} propose a text-to-triple translation method that given a text and a set of entities returns the existing relations. For EE, \citet{lyu-etal-2021-zero} propose, similar to us, the use of an entailment model, but in their case the input sentence is split in clauses according to the output of a Semantic Role Labelling system. In order to compare their results with ours, we only use the event types, not the trigger information\footnote{Output kindly provided by the authors.}. The results from our system can be seen as an ablation where we do not make use of any SRL preprocessing. For EAE, \citet{liu-etal-2020-event} perform zero-shot EAE by recasting the task as QA. Some of these approaches also optimize a threshold on development data, although it is not always clear. We show that our toolkit with default threshold obtains excellent results despite being an all-in-one method.



\section{Discussion} \label{sec:discussion}

\paragraph{Towards post-editing on IE.} Our internal evaluation suggest that verbalizing-while-defining workflow can have similar impact as post-editing machine translated text, where human translators obtain quality translations with less effort~\cite{toral2018post}. The idea of this new framework will bring down the effort required to create larger and higher quality datasets. Current IE system are subject to a predefined schema and are useless to classify new types of entities, relations and events. The use interface of ZS4IE brings to the annotators the opportunity of defining the schema interactively and manually annotating the dataset with the help of the entailment model. In the future we would like to use the manual annotations to fine-tune the TE model, which would further improve the performance, as shown by the excellent few-shot results of \citet{sainz-etal-2021-label}.


\paragraph{Implicit events extraction.} During the development of the EE verbalizations we found out that the entailment model is prone to predict implicit events that are implied by other events. For example, an event type of  \textsc{Justice:Jail} implies an event of \textsc{Justice:Convict} where as the same time it implies event type of \textsc{Justice:Trial-Hearing}. As the entailment models are not specifically trained for a particular IE task (e.g. EE) they are not limited to the  extraction of \textbf{explicit} mentions of types (e.g. event types) annotated in the dataset. We think that this phenomenon might have penalized the RoBERTa* model on the EE task, as ACE dataset only contains annotations of explicit events. On the contrary, rather than a limitation of our approach, we believe that this is a positive feature that can be exploited by the users. 


\section{Conclusions}

The ZS4IE toolkit allows a novice user to model complex IE schemas, curating simple yet effective templates for a target schema with new types of entities, relations, events, and event arguments. Empirical validation showed that reformulating the IE tasks as an entailment problem is easy and effective, as spending only 5-15 minutes per type allows to  achieve very strong zero-shot performance. 
ZS4IE brings to the users the opportunity of defining the desired schema on the fly. In addition it allows to annotate examples, similar to post editing MT output.
Rather than being a finalized toolkit, we envision several exciting directions, such as including further NLP tasks, allowing the user to select custom pre-processing steps for candidate generation and allowing the user to interactively improve the system annotating examples that are used to fine-tune the TE model.

More generally, we would like to extend the inference capability of our models, perhaps acquired from other tasks or schemas \cite{sainz-etal-2022-textual}, in a research avenue where entailment and task performance improve in tandem.  

\section*{Acknowledgements}
Oscar is funded by a PhD grant from the Basque Government (PRE\_2020\_1\_0246). This work is based upon work partially supported via the IARPA  BETTER Program contract No. 2019-19051600006 (ODNI, IARPA), and by the Basque Government (IXA excellence research group IT1343-19).

\bibliography{anthology,custom}
\bibliographystyle{acl_natbib}

\appendix

\section{User Interface}
\label{sec:ui}

We present more details on our user interface (UI) in this section. Our system supports all 4 IE tasks into a single integrated interface. 

\paragraph{Template development. }

\begin{figure*}
\begin{subfigure}{1.0\textwidth}
  \centering
  \includegraphics[width=0.8\linewidth]{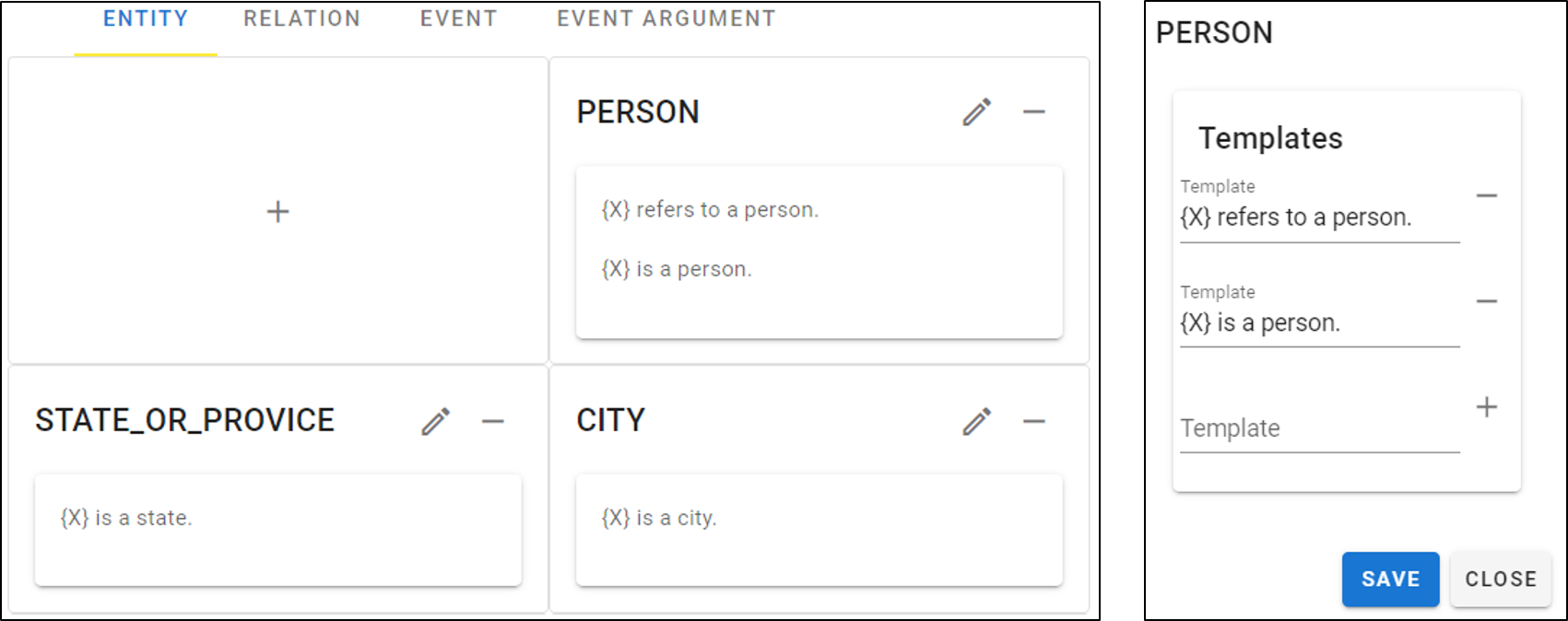}  
  \vspace{-2mm}
  \caption{NER}
  \label{fig:ui3}
\end{subfigure}


\begin{subfigure}{1.0\textwidth}
  \centering
  \includegraphics[width=0.8\linewidth]{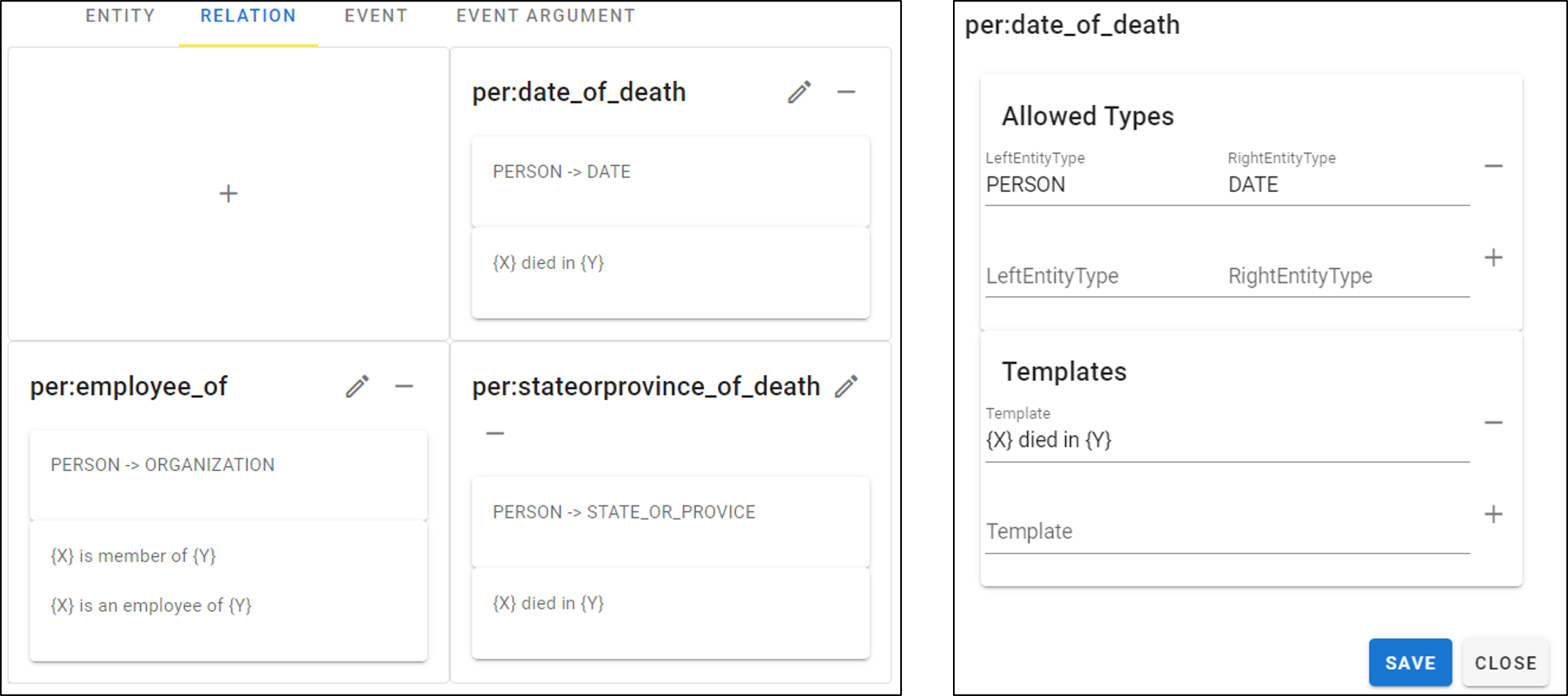}  
   \vspace{-2mm}
  \caption{Relation extraction}
  \label{fig:ui4}
\end{subfigure}

\begin{subfigure}{1.0\textwidth}
  \centering
  \includegraphics[width=0.8\linewidth]{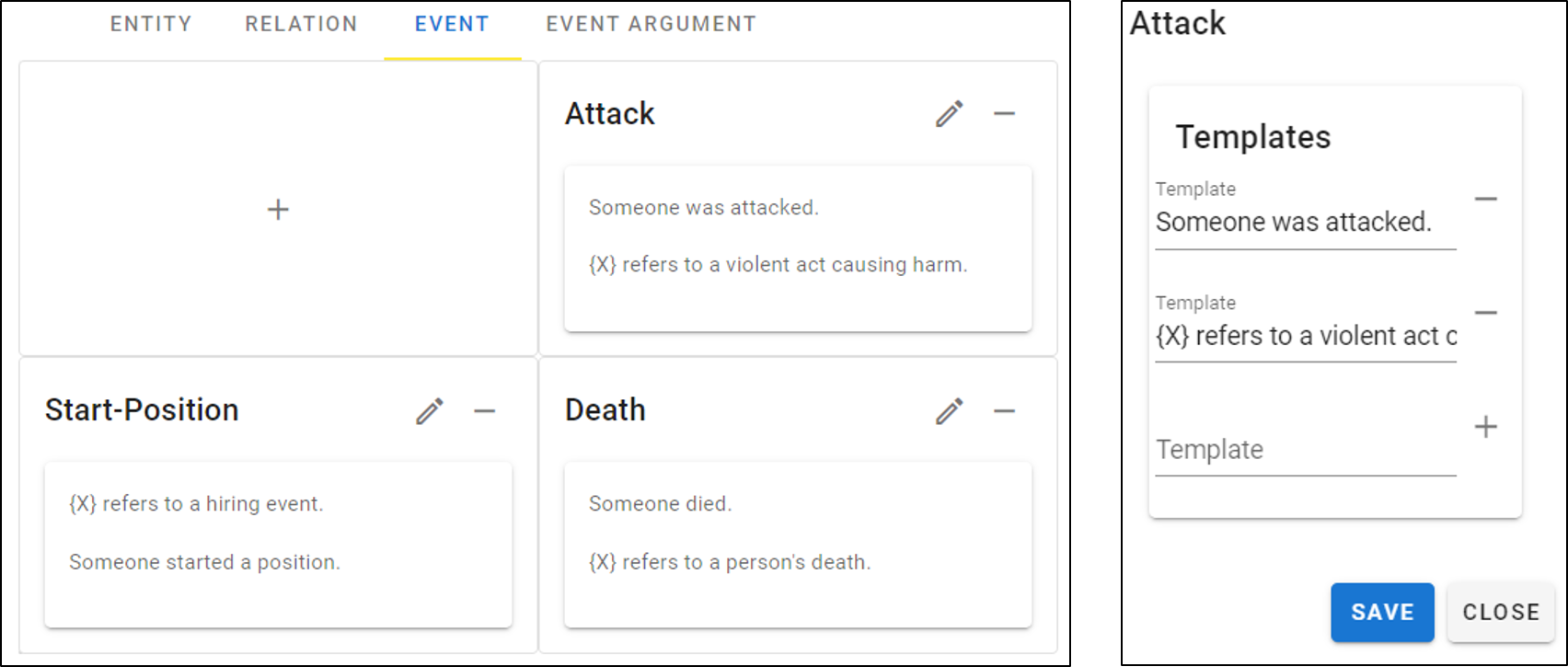}
   \vspace{-2mm}
  \caption{Event extraction}
  \label{fig:ui5}
\end{subfigure}


\begin{subfigure}{1.0\textwidth}
  \centering
  \includegraphics[width=0.8\linewidth]{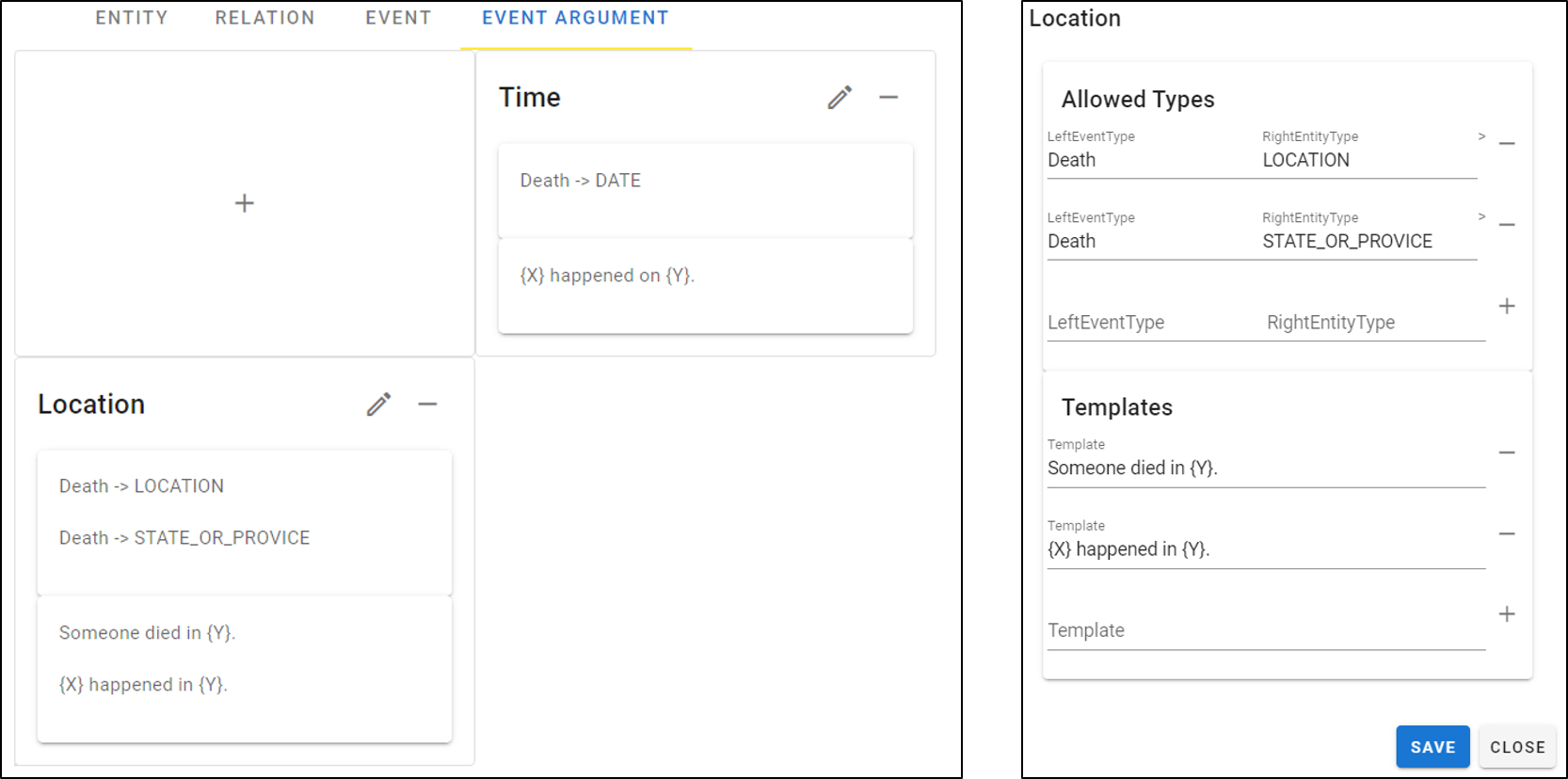}
   \vspace{-2mm}
  \caption{Event argument extraction}
  \label{fig:ui6}
\end{subfigure}
  \vspace{-3mm}
\caption{The UI for developing templates for the 4 IE tasks. For each task, we show the overall UI on the left, and the pop-up window for adding a new entity type PERSON on the right.}
\label{fig:ui7}
\end{figure*}

Figure ~\ref{fig:ui3} shows the main template development UI, in which each tab on the top represents one of the entity, relation, event, and event argument tasks. The user switch between tasks by simply clicking on a different tab (the tabs for the other 3 tasks are shown in Figure~\ref{fig:ui4}, ~\ref{fig:ui5}, and ~\ref{fig:ui6}, respectively). 

Take the NER task as an example (Figure~\ref{fig:ui3}), it shows an overview of all entity types along with the templates defined for each type (e.g., ``{X} is a person'' for the type \textsc{Person}, in which ``{X}'' is a placeholder that can be replaced with a noun phrase ``New York City''). If the user clicks on the edit button (the pen-shaped button), the pop-up window for adding a new entity type (the right-hand side figure in Figure~\ref{fig:ui3}) shows up. The user can add a template by clicking on "+" sign, and then input the template to the left (the user can repeat this several times to add more templates). The user can remove a template by clicking on "-". The user can also click on the big "+" card to the left to add a new entity type. 

Template development for the relation extraction task is similar to NER, except for two differences: first, as shown in Figure~\ref{fig:ui4} (right), we can further add a set of "allowed type" pairs, that are the set of entity pairs each relation is defined over. For example, the ``per:date\_of\_death'' relation is only valid between a pair of \textsc{Person} and \textsc{Date} mentions. Our UI allows the user to specify the ``LeftEntityType'' (left entity type) and the ``RightEntityType'' (right entity type) for each relation type under "allowed type". These type constraints are show on the top box for each relation card on the left figure in Figure~\ref{fig:ui4} (e.g., "\textsc{Person}->\textsc{Date}" under ``per:date\_of\_death''). Second, a relation involves a pair of entity mentions. Therefore, each pattern has two placeholders, ``{X}'' and ``{Y}'', which can be replaced with two entity candidates that are likely to participate in the relationship.

Template development for the event extraction task (Figure~\ref{fig:ui5}) is also similar to NER, except that the template may not contain any trigger. For example, ``Someone died'' is a template for the ``Death'' event (Figure~\ref{fig:ui5}). This template would allow the TE approach to classify whether an extent (e.g., a sentence) expresses a type of event. 

Template development for the event argument extraction task (Figure~\ref{fig:ui6}) is similar to relation extraction, except that the template can include either two placeholders ``{X}'' and ``{Y}'' in which ``{X}'' is an event trigger and ``{Y}'' is an event argument candidate filler (an entity), or only one placeholder ``{Y}'' which is the event argument candidate filler. The later would require the template to implicitly describes the event type as well (for example, ``Someone died in {Y}'' for the \textsc{Location} event argument role in Figure~\ref{fig:ui6}).

\begin{figure*}[h]
    \centering
    \resizebox{\linewidth}{!}{
        \includegraphics{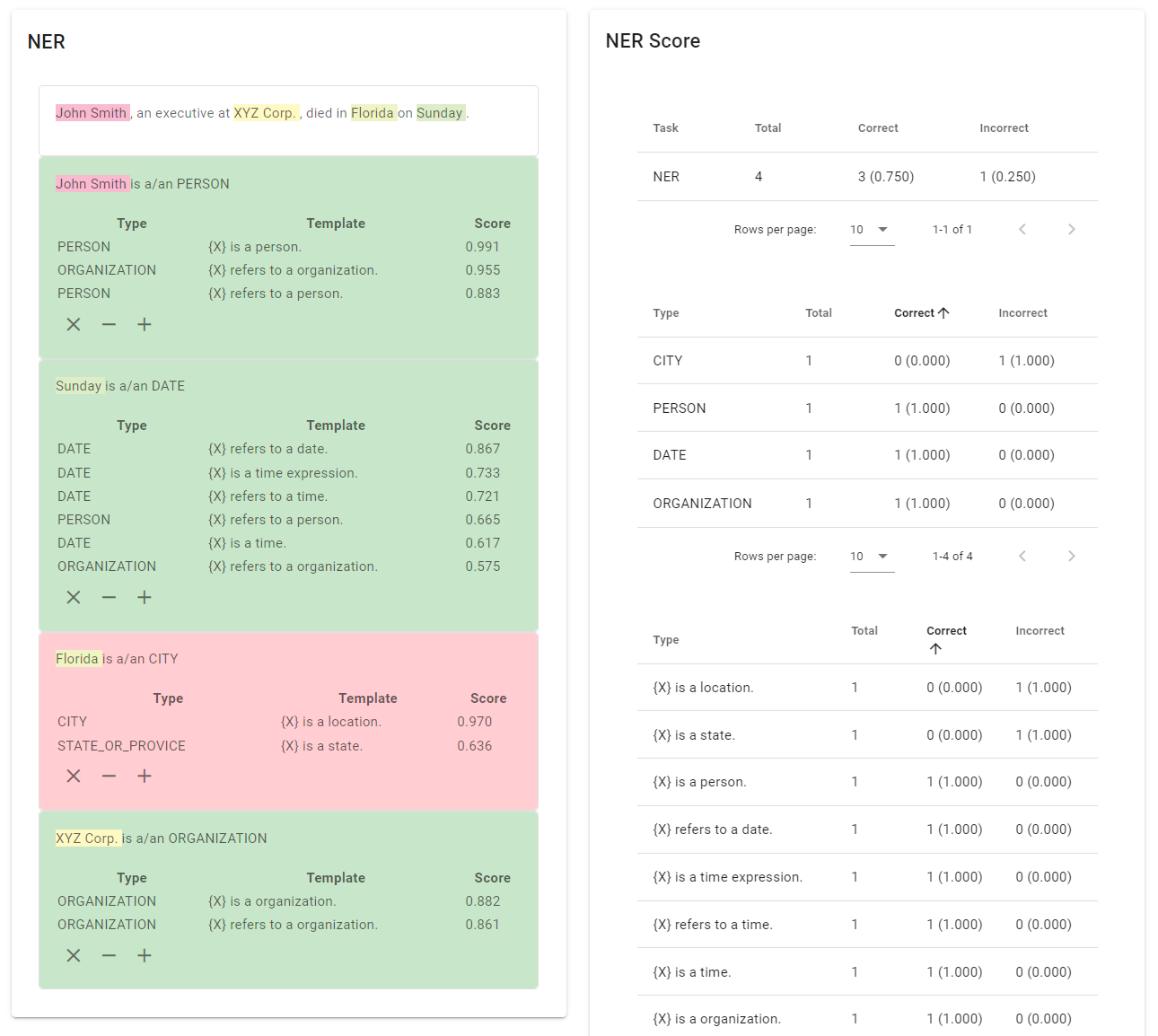}
    }
    \caption{The UI for displaying NER extraction outputs (left) and automatically calculated metrics (right). The left-hand side shows the named entities extracted from an user-input sentence (shown on the top). The user can click on "+" and "-" sign next to each extraction to label its correctness. The right-hand side shows the total number of extracted named entities (total), the number of correct and incorrect extractions (the accuracy number is also shown in the parenthesis next to ``correct'') for the overall task, each type, and each pattern. These metrics are calculated based on the set of user labels.
    }
    \label{fig:ui8}
\end{figure*}

\paragraph{Template validation.} We developed an interactive workflow to allow the user to quickly develop templates and validate their effectiveness in our TE-based framework. To support this workflow, our UI allows the user to run inference over any free text supplied by the user herself/himself. For simplicity, we omit the UI where we allow the user type in free text. We show the UI that displays the extraction output on the free text, that also allows the user to label the correcness of the extractions. Based on those labeled examples, the UI also automatically calculate a few metrics to help the user to find the effectiveness of the templates curated so far.

Figure~\ref{fig:ui8} shows the UI for displaying NER extraction outputs (left) and automatically calculated metrics (right). Taken the user-supplied sentence ``John Smith, an executive at XYZ Corp., died in Florida on Sunday'' as input, the UI on the left-hand side shows the extracted named entities. It shows extractions such as ``John Smith is a/an PERSON'', ``Sunday is a/an DATE'', and so on. To provide rationale for each extraction, it displays a rank list of possible entity types, the template led to that type, along with the entailment probability produced by the pre-trained TE model. The user can click on "+" and "-" sign next to each extraction to label its correctness. In Figure~\ref{fig:ui8}, all extractions are green (labeled by the user as correct) except that ``Florida is a/an CITY'' is in red (labeled as incorrect by the user). Based on these user-labeled extractions, the system calculated a number of metrics to facilitate template validation: the total number of extracted named entities (shown under ``total''), the number of correct and incorrect extractions under ``correct'' and ``incorrect'', respectively (the accuracy number is also shown in the parenthesis next to ``correct'') for the overall task, each type, and each pattern. The right-hand side UI in Figure~\ref{fig:ui8} displays these metrics, and allows the user to sort patterns/types by each of the metric. The user can quickly identify some templates are low-precision (e.g., ``{X} is a location'' for the entity type \textsc{City}), and can revise them to improve precision.

\begin{figure}
\begin{subfigure}{0.47\textwidth}
  \centering
  \includegraphics[width=1.0\linewidth]{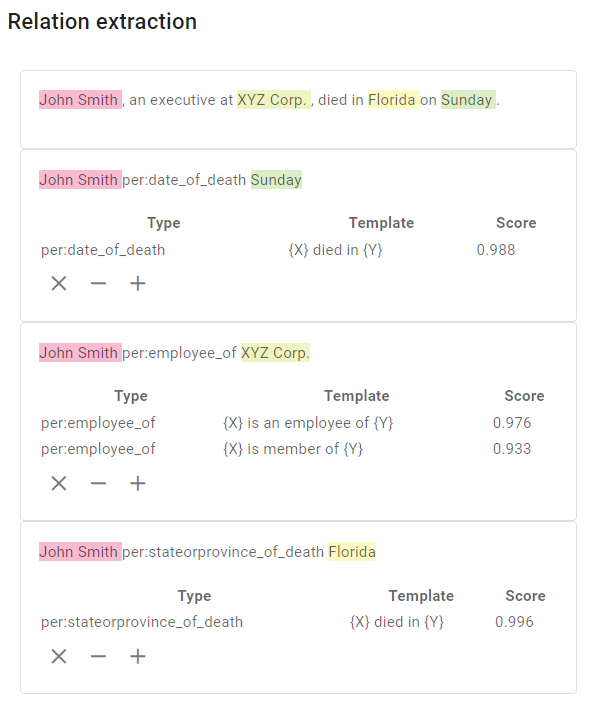}  
  \vspace{-7mm}
  \caption{Relation extraction}
  \label{fig:ui12}
\end{subfigure}


\begin{subfigure}{0.47\textwidth}
  \centering
  \includegraphics[width=1.0\linewidth]{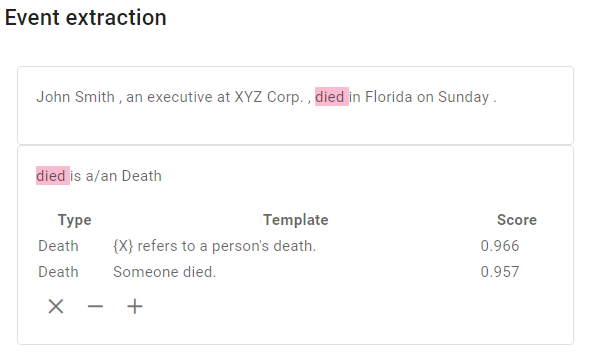}
    \vspace{-7mm}
  \caption{Event extraction}
  \label{fig:ui13}
\end{subfigure}


\begin{subfigure}{0.47\textwidth}
  \centering
  \includegraphics[width=1.0\linewidth]{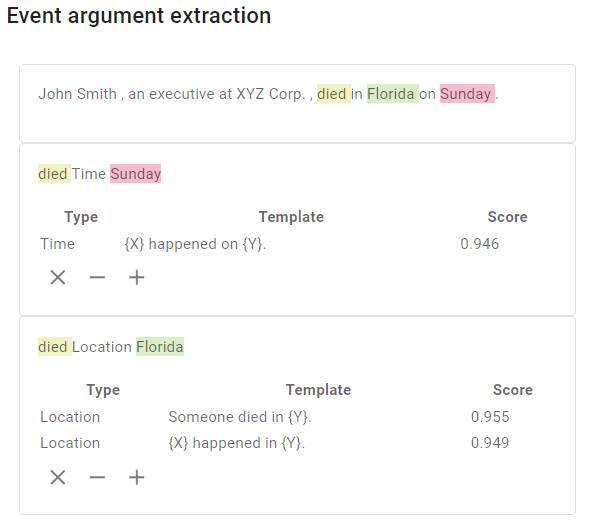}
   \vspace{-7mm}
  \caption{Event argument extraction}
  \label{fig:ui14}
\end{subfigure}
 \vspace{-1mm}
\caption{The UI for displaying extractions for relation extraction, event extraction, and event argument extraction, respectively. The user an click on the "+" or "-" sign next to each extraction to label the extraction as correct or incorrect.}
\label{fig:ui15}
\end{figure}

Figure~\ref{fig:ui12}, ~\ref{fig:ui13}, and ~\ref{fig:ui14} shows the UI for displaying extraction results for the relation extraction, event extraction, and event argument extraction, respectively. Similar to the NER task. Similarly, our system also includes metric board (the metrics above) for the other 3 IE tasks. To view the metric boards for these tasks, please refer to our demonstration video.

\end{document}